\documentclass{article}

     \usepackage[preprint]{neurips_2018}

\usepackage[utf8]{inputenc} 
\usepackage[T1]{fontenc}    
\usepackage{hyperref}       
\usepackage{url}            
\usepackage{booktabs}       
\usepackage{amsfonts}       
\usepackage{nicefrac}       
\usepackage{microtype}      

\usepackage{xcolor}
\usepackage{graphicx}
\usepackage{subcaption}
\usepackage{algorithm}
\usepackage{algpseudocode}
\usepackage{amsmath}
\usepackage{adjustbox}

\title{Fair Credit Scorer through Bayesian Approach}

\author{%
  Zhuo Zhao\\
  Department of Applied Mathematics\\
  Johns Hopkins University\\
  \texttt{zzhao62@jhu.edu} \\
}

\newcommand{\ziyan}[1]{\textcolor{orange}{{\textbf{Ziyan: }\emph{#1}}}}

\begin{document}

\maketitle

\begin{abstract}
  Machine learning currently plays an increasingly important role in people's lives in areas such as credit scoring, auto-driving, disease diagnosing, and insurance quoting. However, in many of these areas, machine learning models have performed unfair behaviors against some sub-populations, such as some particular groups of race, sex, and age. These unfair behaviors can be on account of the pre-existing bias in the training dataset due to historical and social factors. In this paper, we focus on a real-world application of credit scoring and construct a fair prediction model by introducing latent variables to remove the correlation between protected attributes, such as sex and age, with the observable feature inputs, including house and job. For detailed implementation, we apply Bayesian approaches, including the Markov Chain Monte Carlo simulation, to estimate our proposed fair model.
\end{abstract}

\section{Introduction}

Nowadays, Machine Learning methods are used to automate decisions in a variety of areas, including determining credit scores \cite{nanni2009experimental}, classifying tumor components from MRI images \cite{lundervold2019overview}, detecting pedestrians on the road \cite{dollar2011pedestrian}, and understanding natural languages \cite{goldberg2014word2vec}, etc. However, machine learning methods are heavily dependent on data \cite{mitchell1997machine} and this data-dependent nature makes the learned models sensitive to the latent bias existing in the training datasets \cite{mehrabi2021survey}. Thus, the final decisions made by the learned models are unfairly biased against certain sub-populations, differentiated by some \textit{sensitive/protected attributes}, such as race, sex, or age, etc. For example, cameras sometimes fail to recognize whether Asian blink their eyes \cite{camera_pages} and the beauty pageant judged by AI would prefer light skin \cite{the_guardian_2016_AI_judge}. However, we would expect AI to give the same decision independent from the protected attributes and thus we concern about the \textit{fairness} of machine learning methods \cite{mehrabi2021survey}.

In this paper, we focus on constructing fair machine learning models to predict the credit score, with using the German Credit Risk dataset \cite{german_credit_risk} (Sec. \ref{sec:dataset}). The goal is to predict the credit score based on some \textit{observable variables}, including housing and job information. However, this personal financial information, such as income, housing, and saving, are usually highly correlated to gender and age due to historical and social reasons \cite{rennison2003nonlethal}. Therefore, it is necessary to learn an effective model to filter the prediction bias against sex and age, caused by the latent correlation between these observable variables and the protected attributes. In detail, we analyze and compare from the fairness perspective across the full model \cite{montgomery2021introduction}, unaware model \cite{dwork2012fairness}, and fair model based on causals and counterfactuals \cite{kusner2017counterfactual} (Sec. \ref{sec:methods}). Then, we apply the Markov Chain Monte Carlo (MCMC) simulation \cite{mooney1997monte} and the Gibbs' sampling \cite{gelfand2000gibbs} to solve the corresponding parameters in these models and evaluate the performances (Sec. \ref{sec:exp} and Sec. \ref{sec:result}).
\section{Related work}

\textbf{Fairness}. Many recent works (\cite{calders2010three, bolukbasi2016man, dwork2012fairness, hardt2016equality, joseph2016rawlsian, kusner2017counterfactual}) have been focusing on fairness in machine learning algorithms. \cite{bolukbasi2016man} pointed out that there is a risk of amplifying the bias introduced from the dataset, if using machine learning algorithms without taking effects to handle the pre-existing bias. For example, in the word embedding, learned over Google News with pre-existing gender stereotypes, the gender-neutral words widely spread along a latent embedding direction capturing gender difference,  such as "receptionist" falling far along the direction related to "female" \cite{bolukbasi2016man}. \cite{calders2010three} modifies the Naive Bayes classifier by adding independence restriction toward sensitive attributes. \cite{dwork2012fairness} proposes a task-specific metric to evaluate the similarity between individuals relative to the classification task and optimizes over the proposed metric with the goal that similar individuals are treated similarly in the classification task. \cite{kusner2017counterfactual} focuses on causal inference and counterfactual, with introducing the latent confounding variables, which are related to the observable variables but independent from the protected attributes. Our work is based on the \cite{kusner2017counterfactual} idea to construct a fair prediction model over the German Credit Risk dataset \cite{german_credit_risk}.

\section{Dataset}
\label{sec:dataset}
We consider the Kaggle German Credit Risk dataset \cite{german_credit_risk} to analyze and compare different types of unfair models and our method for constructing a fair model using Bayesian approaches. In this dataset, each entry represents a person who takes credit from a bank. The objective is to predict the credit amount of a person based on his/her attributes. "Sex" and "age" are the sensitive/protected attributes related to the bias during training and prediction in the unfairness problem. Feature "job" is a binary variable representing whether a person has a job or not. Feature "house" is a binary variable that indicates whether or not a person owns a house. The "credit amount" is our prediction target. 

\begin{figure}[h]
    \centering
    \includegraphics[width=1\textwidth]{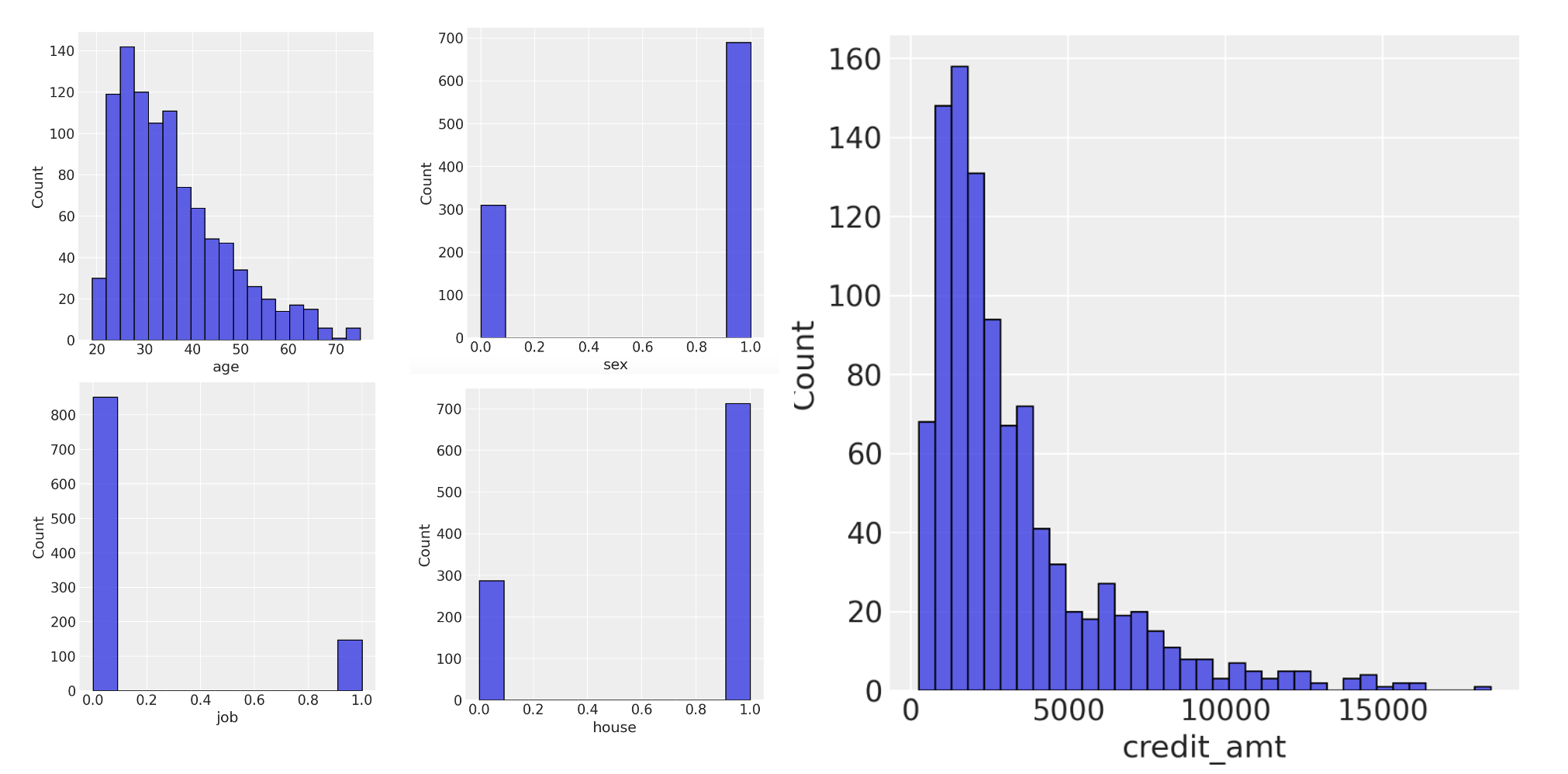}
    \caption{Distribution of features in the German Credit Risk dataset \cite{german_credit_risk}. "Age" and "sex" are the sensitive / protected attributes. "Job" and "house" are the observable variables. "Credit amount" is the prediction target.}
    \label{fig:dataset info}
\end{figure}

\begin{figure}[h]
    \centering
    \includegraphics[width=0.6\textwidth]{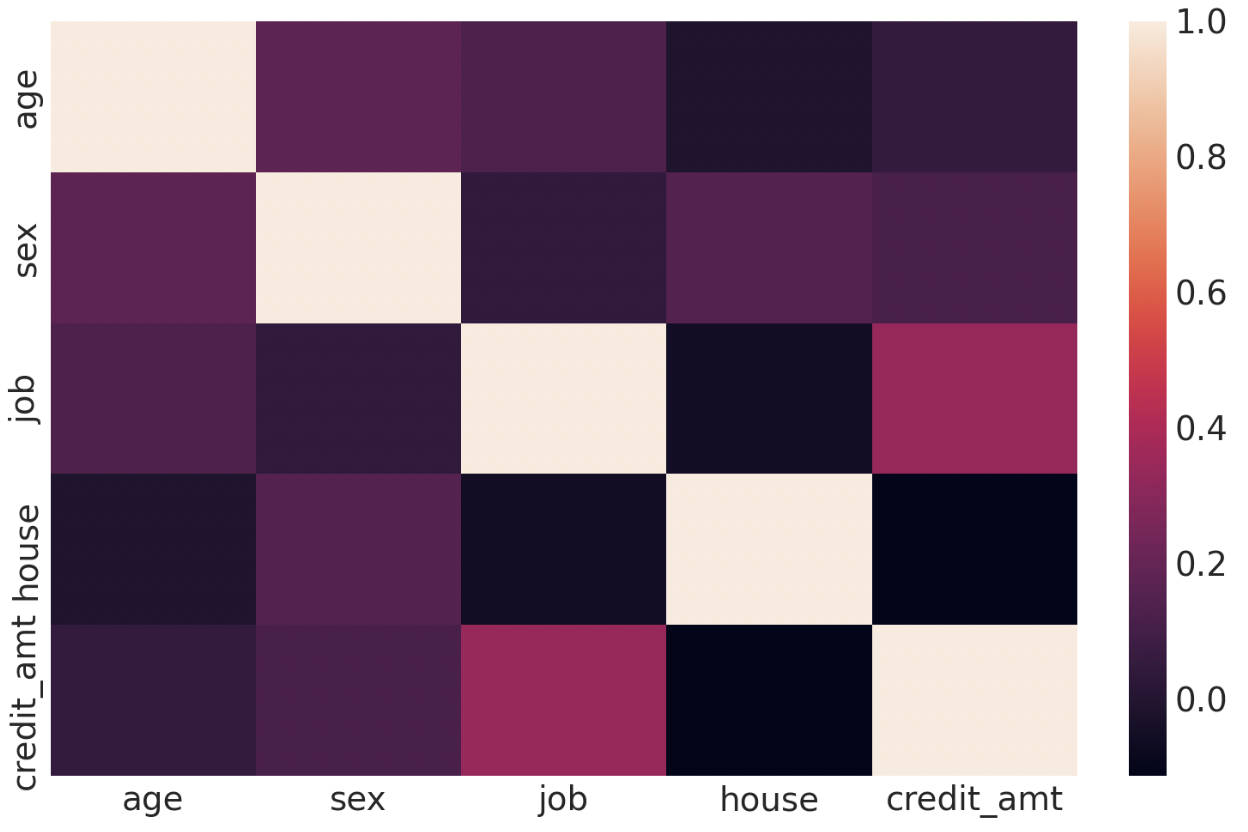}
    \caption{Illustration of the covariance matrix between all the input features and the prediction target. Here, we observe a high correlation from "age" and "sex" (the sensitive/protected attributes) to "job" and "house" (the observable variables).}
    \label{fig:conv matrix}
\end{figure}

The dataset is composed of 1000 records. We randomly pick 800 records for training and 200 records for testing. Figure \ref{fig:dataset info} shows the detailed distributions of all these features in the whole dataset. In Figure \ref{fig:conv matrix}, we illustrate the covariance between all the input features and the prediction target. We can observe a high correlation from the sensitive / protected attributes, i.e. "sex" and "age", to the "job" and "house". Thus, it is necessary to consider the issue of fairness when constructing a prediction model over "job" and "house".

\section{Methods}
\label{sec:methods}
\begin{figure}[ht!]
    \centering
    \includegraphics[width=1\textwidth]{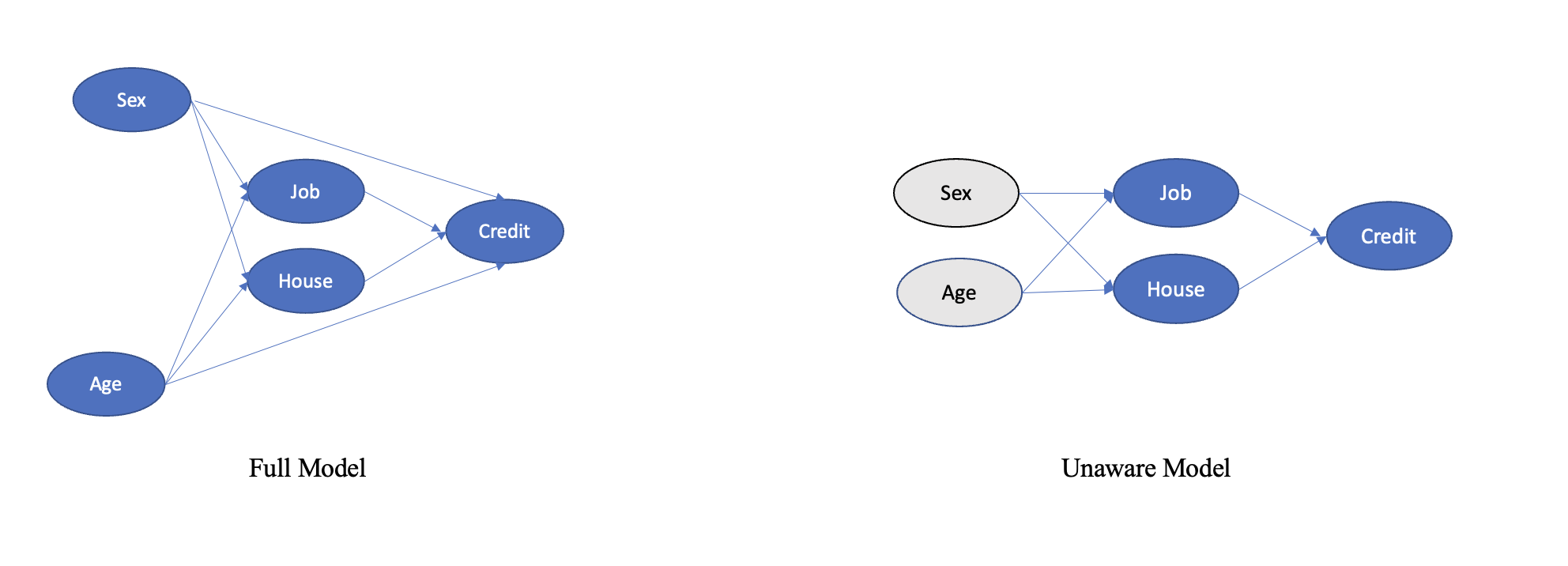}
    \caption{Two types of unfair models. Left: full model, which builds regression over all possible attributes without the consideration of fairness. Right: unaware model, which excludes sensitive/protected attributes, i.e. sex and age in our case.}
    \label{fig:unfair model}
\end{figure}

\textbf{Full Model:} The full model \cite{montgomery2021introduction} completely ignores fairness issues and includes sensitive variables like sex and age in the learning process. It is easy to understand that the full model is unfair because the predictions depend on sex and age. Figure \ref{fig:unfair model} presents the directed acyclic graph (DAG) of the full model. In the full model, all the features are assumed to be connected. \\ \\
\textbf{Unaware Model:} The unaware model \cite{dwork2012fairness} does not use sensitive variables in the learning and prediction process, but it is still unfair. Even though the sensitive variables do not influence the target directly in the learning and prediction processes, it still has an indirect impact on the target through the non-sensitive variables. In our example, to predict a person's credit amount, sex may influence whether a person can get a job. The job attribute still preserves the information of sex. Simply ignoring the sex attribute will not fully eliminate its impact on the predictions. Figure \ref{fig:unfair model} presents the DAG of an unaware model. The attributes under the grey circles are unobserved. In the unaware model, sex and age are not directly connected with the credit amount, but they are connected with job and house. It is still unfair because the change of sex and age will change the status of job and house, and thus influence the credit amount predictions.

\begin{figure}[ht!]
    \centering
    \includegraphics[width=1\textwidth]{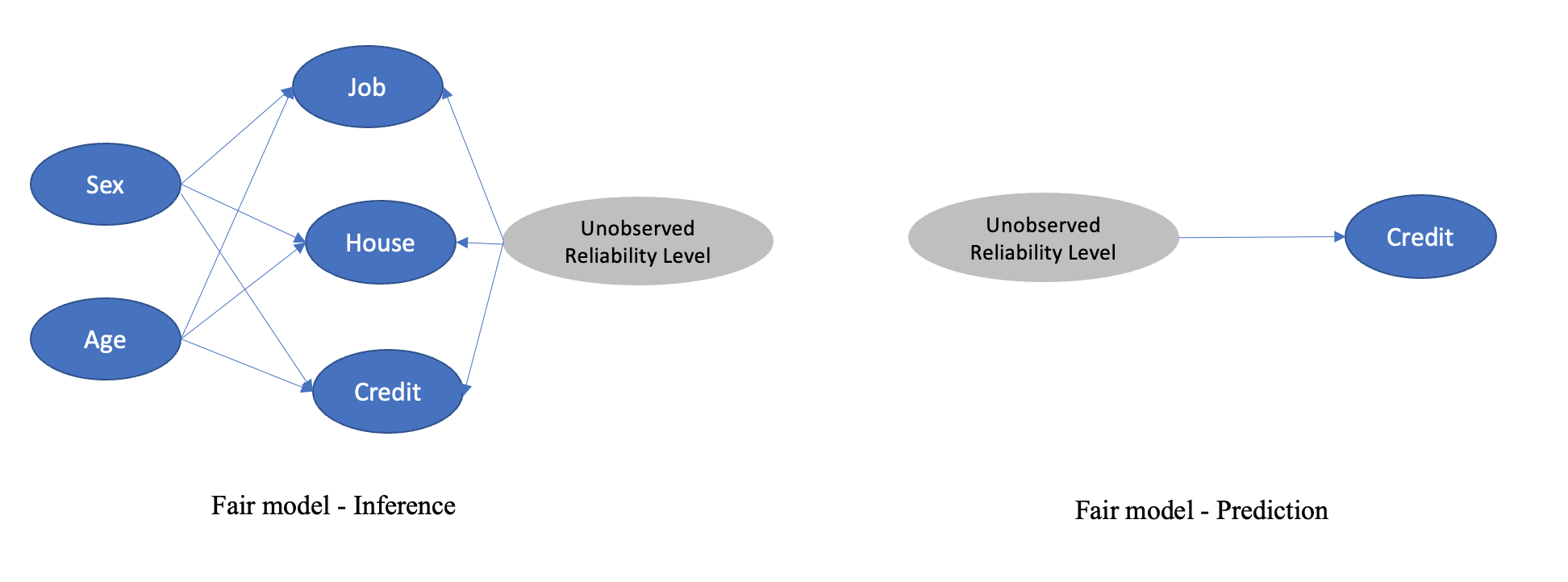}
    \caption{DAG of the fair model. Here, we introduce the latent confounding variable "unobserved reliability level", which is independent to "sex" and "age" (the sensitive/protected attributes) but related to "job", "house", and "credit amount". Left: during the inference stage, we estimate this latent "reliability" feature with Bayesian approaches. Right: during the prediction stage, we only use this inferred "reliability" feature to predict the "credit amount".}
    \label{fig:fair model}
\end{figure}

\textbf{Fair Model: } In order to build a fair model, we need to find a proxy variable that is independent of sensitive variables but still preserves the information in the credit amount prediction \cite{kusner2017counterfactual}. We can introduce the concept of latent \textit{confounding variable} to resolve this issue. The confounding variable is a variable that influences both the independent variable and dependent variables. In our fair model, we assume that there is an unobserved confounder $C$ that reflects how reliable a person is in paying back the loan. The confounder should be independent of the sensitive variables to make the model fair. Figure \ref{fig:fair model} shows the DAG of the fair model structure. In the inference stage, we assume that job, house, and credit amount are confounded by the unobserved reliability level $C$ and $C$ is independent of sex and age. The reason is that sex and age can neither determine nor be related to how reliable a person is in paying back loans. Meanwhile, reliability is co-related to a person's job performance, housing situation, and also credit amount. Then, in the prediction stage, we only use the inferred $C$ as our feature to predict the credit amount. In this way, the predicting process does not contain any information about sex or age, and thus this procedure is an effective, fair learning algorithm in our scenario.

\section{Experiments}
\label{sec:exp}
We can represent the DAG of the fair model in a probabilistic way. We sample the two binary variables, job and house from two Bernoulli distributions and sample the confounder from the normal distribution. In the meantime, we choose the Poisson distribution as a prior for the credit amount. Our choices of priors correspond to the nature of the data. The job and house features are binary. And the credit amount is a positive attribute with a shape alike the Poisson distribution. The probabilistic model can be written as:
\begin{equation}
    Job \sim Bernoulli(logit(b_j + Sex \times \beta_{j,s} + Age \times \beta_{j,a} + C \times \beta_{j,c}))
    \label{eq:job}
\end{equation}
\begin{equation}
    House \sim Bernoulli(logit(b_h + Sex \times \beta_{h,s} + Age \times \beta_{h,a} + C \times \beta_{h,c}))
    \label{eq:house}
\end{equation}
\begin{equation}
    Credit \sim Poisson(Exp(Sex \times \beta_{c,s} + Age \times \beta_{c,a} + C \times \beta_{c,c}))
    \label{eq:credit}
\end{equation}
\begin{equation}
    {C \sim Normal(0, 1)}  \text{\quad where \quad} C \perp Sex, \text{\quad} C \perp Age 
    \label{eq:reliability}
\end{equation}
The parameters we need to find are in the set $\Theta = \{\beta_{m, n}, b_m\}$ where $m = j, h$ and $n =s, a, c$. We assume that these parameters are sampled from the normal distributions:
\begin{equation}
    \beta_{m,n} \sim N(0,1)
\end{equation}
\begin{equation}
    b_m \sim N(0, 1)
\end{equation}
We implement the Metropolis–Hastings algorithm to infer the probabilistic model. M-H algorithm \cite{hastings1970monte} is a Markov Chain Monte Carlo (MCMC) method for obtaining a sequence of random samples from a probability distribution from which direct sampling is difficult. Algorithm \ref{alg:cap} explains how to infer the reliability level $C$.
\begin{algorithm}
\caption{Infer C by Metropolis–Hastings}\label{alg:cap}
\begin{algorithmic}
\For{$i=1$ to $N$}
\State Choose $J(C_i^* | C_i^{(s)}) = uniform(C_i^{(s)} - \delta, C_i^{(s)} + \delta)$;
\State Set an initial state $C_i^{0}$;
\For{$s=1$ to $5000$} 
\State Sample $C_i^* \sim J(C_i^* | C_i^{(s)})$;
\State Compute the acceptance ratio $r = \frac{p(C_i^*\mid y)}{p(C_i^{(s)}\mid y)}=\frac{p(y \mid C_i^*)p(C_i^*)}{p(y \mid C_i^{(s)})p(C_i^{(s)})}$;
\State sample $u \sim uniform(0, 1)$;
\If{u < r};
\State $C_i^{(s+1)} = C_i^*$;
\Else
\State $C_i^{(s+1)} = C_i^{(s)}$;
\EndIf
\EndFor
\EndFor
\end{algorithmic}
\end{algorithm}

Once we obtain the posteriors of the inferred reliability level $C$, we can fit a new model using kernel $g(.)$ based on the $C$ in the prediction stage. In our experiment, since there is a nonlinear relationship between credit amount and "Reliability Level" in our inference stage setup (Poisson), we decide to use random-forest as the kernel function $g(.)$ in our second stage prediction. 
\begin{equation}
    Credit \sim g(C)
\end{equation}

\section{Results}
\label{sec:result}
In this section, we provide experimental results and a discussion of the MCMC process performance. Specifically, in Sec. \ref{subsec:mcmc_performance}, we firstly present the MCMC estimation result and the convergence analysis on the fair model's latent confounding variable $C$ and parameters. Then, we compare the prediction and fairness performance across the three types of models in Sec. \ref{subsec:cross_comparison}.
\subsection{Fair model's MCMC performance:}
\label{subsec:mcmc_performance}
\begin{figure}[ht!]
    \centering
    \includegraphics[width=1\textwidth]{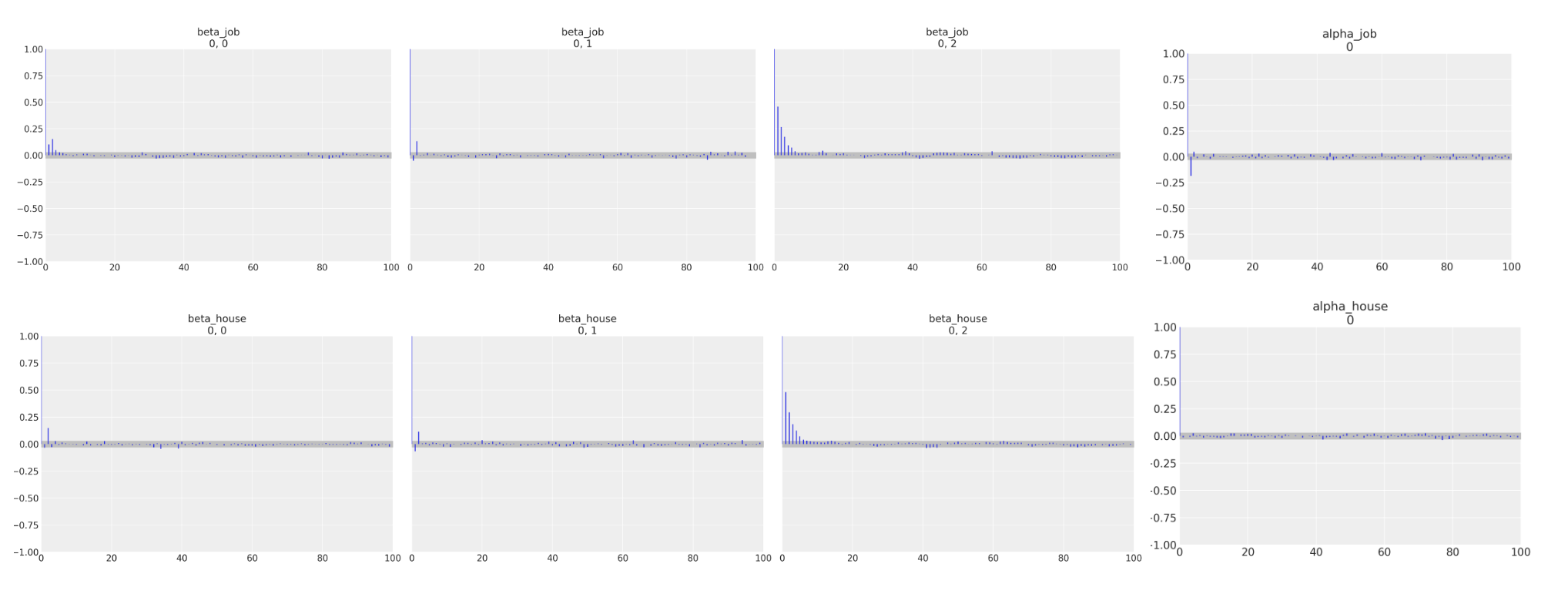}
    \caption{Auto-correlation plots of parameters in Eq. \ref{eq:job} and Eq. \ref{eq:house} throughout the MCMC process. "alpha" refers to the constant offset term $b$ in the equations.}
    \label{fig:auto_corr}
\end{figure}
\begin{figure}[]
    \centering
    \includegraphics[width=1\textwidth]{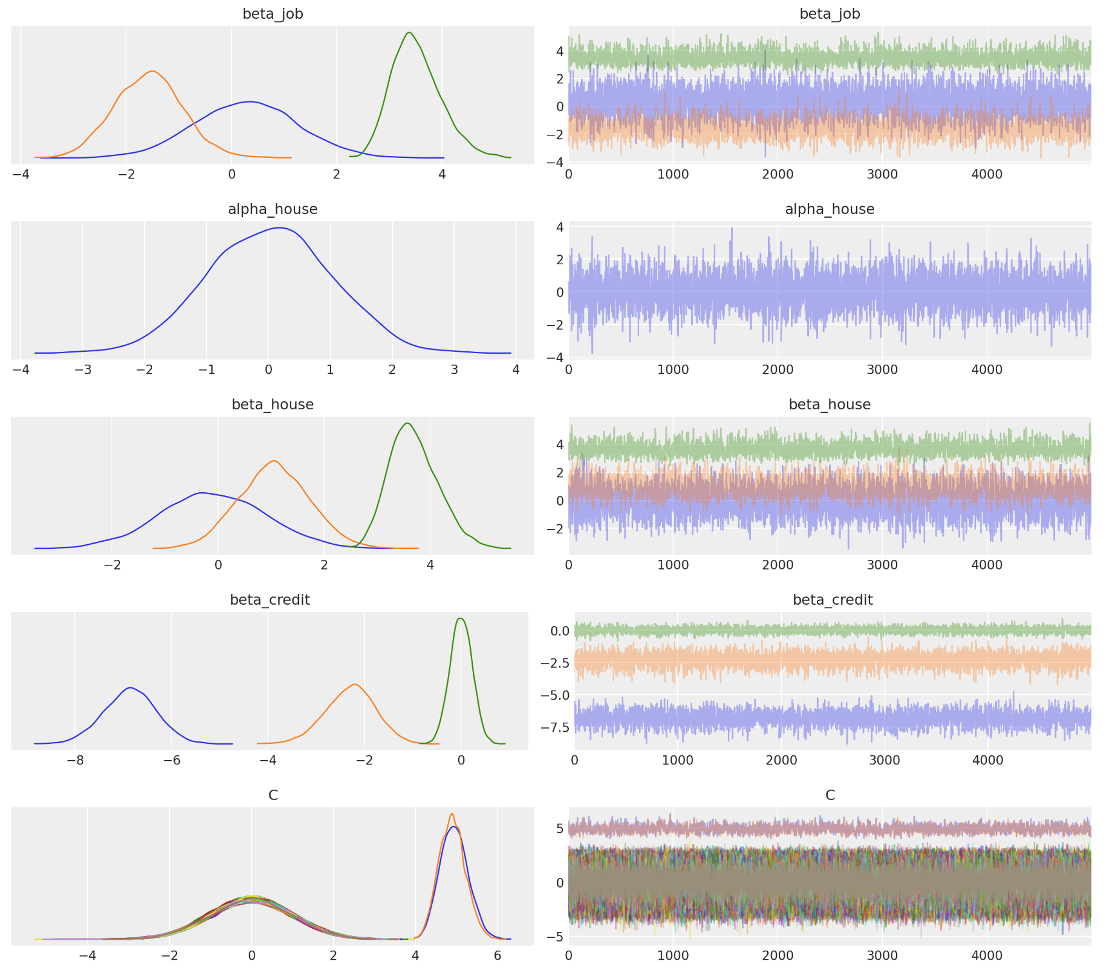}
    \caption{Posterior estimation (left column) and trace plot (right column) of parameters in Eq. \ref{eq:job} and Eq. \ref{eq:house} throughout the MCMC process. "alpha" refers to the constant offset term $b$ in the equations.}
    \label{fig:trace_plot}
\end{figure}
In Figure \ref{fig:auto_corr}, we illustrate the auto-correlation plot of the model's parameters in Eq. \ref{eq:job} and Eq. \ref{eq:house}. We observe a clear decrease in auto-correlation throughout the MCMC process. Thus, this is an efficient MCMC process that leads to convergence. Further, in Figure \ref{fig:trace_plot}, we provide the posterior estimation and the trace plot of the fair model parameters throughout the MCMC process. Though we still observe some fluctuations till the end of the process, however, this is reasonable and acceptable. The reason is that we are applying over a real-world dataset, rather than a simulated dataset. Therefore, it is impossible to make our assumed distributions perfectly capture the behavior of the real-world dataset. Then, in Table \ref{tab:ci_param}, we provide the confidence interval over the posterior estimation of the fair model's parameters.

\begin{table}[]
\centering
\begin{tabular}{lrrrrrr}
\hline
{} &   std &    5\% &  median &   95\% &  ess\_bulk &  ess\_tail \\
\hline
\hline
$b_j$      &  1.02 & -1.66 &    0.03 &  1.71 &   4643.63 &   3709.55 \\
$\beta_{j, s}$    &  0.98 & -1.32 &    0.27 &  1.88 &   7128.36 &   3907.04 \\
$\beta_{j, a}$    &  0.65 & -2.64 &   -1.57 & -0.50 &   1502.60 &   2245.58 \\
$\beta_{j, c}$    &  0.47 &  2.82 &    3.46 &  4.36 &   2058.64 &   2494.02 \\
$b_h$    &  1.01 & -1.61 &    0.03 &  1.67 &   5113.35 &   3167.94 \\
$\beta_{h, s}$   &  0.99 & -1.73 &   -0.11 &  1.55 &   5506.02 &   3900.82 \\
$\beta_{h, a}$   &  0.67 & -0.04 &    1.05 &  2.17 &   1625.86 &   2583.99 \\
$\beta_{h, c}$   &  0.46 &  3.00 &    3.65 &  4.50 &   1896.31 &   2939.44 \\
$\beta_{c, s}$  &  0.54 & -7.78 &   -6.85 & -5.98 &   3255.68 &   3206.54 \\
$\beta_{c, a}$  &  0.52 & -3.17 &   -2.26 & -1.46 &   4326.49 &   3800.31 \\
$\beta_{c, c}$  &  0.23 & -0.37 &    0.01 &  0.38 &   4455.72 &   3211.78 \\
\hline
\end{tabular}
    \caption{The confidence intervals of the parameters estimated in Eq. \ref{eq:job} and Eq. \ref{eq:house} through the MCMC process.}
    \label{tab:ci_param}

\end{table}

\subsection{Performance comparison across models:}
In this section, we compare how three distinct models perform while making predictions. In Table \ref{tab:performance}, we present the $R^2$ of three models in both training and testing environments. The full model outperforms the unaware model in both fitting and predicting by including sensitive information. It is surprising to see that the fair model outperforms the other two unfair models with $R^2 = 0.801$ in the training set and $R^2 = 0.768$ in the testing set. It turns out that our fair model does not only resolve the fairness issue but distills the information on the reliability level. The fair model is robust enough to be used to make fair and accurate predictions. 
\label{subsec:cross_comparison}

\begin{table}[]
\centering
\begin{tabular}{|lccc|}
\hline
{$R^2$} &   Full Model &  Unaware Model & Fair Model Random Forest Kernel \\
\hline
Training & 0.597 & 0.466 & 0.801 \\ 
Testing & 0.521 & 0.424 & 0.768 \\ 
\hline
\end{tabular}
    \caption{The $R^2$ of three types of models defined in Sec. \ref{sec:methods}.}
    \label{tab:performance}

\end{table}

\section{Conclusion}
In this paper, we have presented a fair model focusing on predicting the German credit score with considering the job and housing features. Specifically, we introduce the latent confounding variable "reliability level", which is independent of the protected attributes, i.e., "sex" and "age", but related to other observable variables and the prediction goal. For implementation, we apply the MCMC approach to solve for the latent confounding variable and the parameters of the model. Compared with tradition models, our model effectively eliminates the bias related to sex and age and thus achieves a fair prediction of the credit amount. For the future work, we recommend trying different types of assumptions on the distribution for the variables over the German Credit Risk dataset and checking the effects on the choice of distributions over the convergence of the MCMC process and the final prediction.
\bibliographystyle{plainnat}
\bibliography{egbib}

\begin{thebibliography}{20}
\providecommand{\natexlab}[1]{#1}
\providecommand{\url}[1]{\texttt{#1}}
\expandafter\ifx\csname urlstyle\endcsname\relax
  \providecommand{\doi}[1]{doi: #1}\else
  \providecommand{\doi}{doi: \begingroup \urlstyle{rm}\Url}\fi

\bibitem[Bolukbasi et~al.(2016)Bolukbasi, Chang, Zou, Saligrama, and
  Kalai]{bolukbasi2016man}
Tolga Bolukbasi, Kai-Wei Chang, James~Y Zou, Venkatesh Saligrama, and Adam~T
  Kalai.
\newblock Man is to computer programmer as woman is to homemaker? debiasing
  word embeddings.
\newblock \emph{Advances in neural information processing systems}, 29, 2016.

\bibitem[Calders and Verwer(2010)]{calders2010three}
Toon Calders and Sicco Verwer.
\newblock Three naive bayes approaches for discrimination-free classification.
\newblock \emph{Data mining and knowledge discovery}, 21\penalty0 (2):\penalty0
  277--292, 2010.

\bibitem[Dollar et~al.(2011)Dollar, Wojek, Schiele, and
  Perona]{dollar2011pedestrian}
Piotr Dollar, Christian Wojek, Bernt Schiele, and Pietro Perona.
\newblock Pedestrian detection: An evaluation of the state of the art.
\newblock \emph{IEEE transactions on pattern analysis and machine
  intelligence}, 34\penalty0 (4):\penalty0 743--761, 2011.

\bibitem[Dwork et~al.(2012)Dwork, Hardt, Pitassi, Reingold, and
  Zemel]{dwork2012fairness}
Cynthia Dwork, Moritz Hardt, Toniann Pitassi, Omer Reingold, and Richard Zemel.
\newblock Fairness through awareness.
\newblock In \emph{Proceedings of the 3rd innovations in theoretical computer
  science conference}, pages 214--226, 2012.

\bibitem[Gelfand(2000)]{gelfand2000gibbs}
Alan~E Gelfand.
\newblock Gibbs sampling.
\newblock \emph{Journal of the American statistical Association}, 95\penalty0
  (452):\penalty0 1300--1304, 2000.

\bibitem[Goldberg and Levy(2014)]{goldberg2014word2vec}
Yoav Goldberg and Omer Levy.
\newblock word2vec explained: deriving mikolov et al.'s negative-sampling
  word-embedding method.
\newblock \emph{arXiv preprint arXiv:1402.3722}, 2014.

\bibitem[Guardian(2016)]{the_guardian_2016_AI_judge}
The Guardian.
\newblock A beauty contest was judged by ai and the robots didn't like dark
  skin, Sep 2016.
\newblock URL
  \url{https://www.theguardian.com/technology/2016/sep/08/artificial-intelligence-beauty-contest-doesnt-like-black-people}.

\bibitem[Hardt et~al.(2016)Hardt, Price, and Srebro]{hardt2016equality}
Moritz Hardt, Eric Price, and Nati Srebro.
\newblock Equality of opportunity in supervised learning.
\newblock \emph{Advances in neural information processing systems}, 29, 2016.

\bibitem[Hastings(1970)]{hastings1970monte}
W~Keith Hastings.
\newblock Monte carlo sampling methods using markov chains and their
  applications.
\newblock 1970.

\bibitem[Hoffman(2016)]{german_credit_risk}
Donald Hoffman.
\newblock German credit risk, Dec 2016.
\newblock URL \url{https://www.kaggle.com/datasets/uciml/german-credit}.

\bibitem[Joseph et~al.(2016)Joseph, Kearns, Morgenstern, Neel, and
  Roth]{joseph2016rawlsian}
Matthew Joseph, Michael Kearns, Jamie Morgenstern, Seth Neel, and Aaron Roth.
\newblock Rawlsian fairness for machine learning.
\newblock \emph{arXiv preprint arXiv:1610.09559}, 1\penalty0 (2):\penalty0 19,
  2016.

\bibitem[Kusner et~al.(2017)Kusner, Loftus, Russell, and
  Silva]{kusner2017counterfactual}
Matt~J Kusner, Joshua Loftus, Chris Russell, and Ricardo Silva.
\newblock Counterfactual fairness.
\newblock \emph{Advances in neural information processing systems}, 30, 2017.

\bibitem[Lundervold and Lundervold(2019)]{lundervold2019overview}
Alexander~Selvikv{\aa}g Lundervold and Arvid Lundervold.
\newblock An overview of deep learning in medical imaging focusing on mri.
\newblock \emph{Zeitschrift f{\"u}r Medizinische Physik}, 29\penalty0
  (2):\penalty0 102--127, 2019.

\bibitem[Mehrabi et~al.(2021)Mehrabi, Morstatter, Saxena, Lerman, and
  Galstyan]{mehrabi2021survey}
Ninareh Mehrabi, Fred Morstatter, Nripsuta Saxena, Kristina Lerman, and Aram
  Galstyan.
\newblock A survey on bias and fairness in machine learning.
\newblock \emph{ACM Computing Surveys (CSUR)}, 54\penalty0 (6):\penalty0 1--35,
  2021.

\bibitem[Mitchell and Mitchell(1997)]{mitchell1997machine}
Tom~M Mitchell and Tom~M Mitchell.
\newblock \emph{Machine learning}, volume~1.
\newblock McGraw-hill New York, 1997.

\bibitem[Montgomery et~al.(2021)Montgomery, Peck, and
  Vining]{montgomery2021introduction}
Douglas~C Montgomery, Elizabeth~A Peck, and G~Geoffrey Vining.
\newblock \emph{Introduction to linear regression analysis}.
\newblock John Wiley \& Sons, 2021.

\bibitem[Mooney(1997)]{mooney1997monte}
Christopher~Z Mooney.
\newblock \emph{Monte carlo simulation}.
\newblock Number 116. Sage, 1997.

\bibitem[Nanni and Lumini(2009)]{nanni2009experimental}
Loris Nanni and Alessandra Lumini.
\newblock An experimental comparison of ensemble of classifiers for bankruptcy
  prediction and credit scoring.
\newblock \emph{Expert systems with applications}, 36\penalty0 (2):\penalty0
  3028--3033, 2009.

\bibitem[Rennison and Planty(2003)]{rennison2003nonlethal}
Callie Rennison and Mike Planty.
\newblock Nonlethal intimate partner violence: Examining race, gender, and
  income patterns.
\newblock \emph{Violence and victims}, 18\penalty0 (4):\penalty0 433--443,
  2003.

\bibitem[Sharp(2009)]{camera_pages}
Gwen Sharp.
\newblock Nikon camera says asians: People are always blinking - sociological
  images, 2009.
\newblock URL
  \url{https://thesocietypages.org/socimages/2009/05/29/nikon-camera-says-asians-are-always-blinking/}.

\end{thebibliography}
\end{document}